\pdfoutput=1
\documentclass[11pt]{article}

\usepackage{acl}

\usepackage{times}
\usepackage{latexsym}

\usepackage[T1]{fontenc}
\usepackage[utf8]{inputenc}

\usepackage{microtype}

\usepackage[T1]{fontenc}
\usepackage{type1cm}
\usepackage[american]{babel}
\usepackage{url}
\usepackage{xspace}
\usepackage{microtype}

\usepackage[pdftex]{graphicx}
\usepackage{booktabs}

\DeclareGraphicsExtensions{.pdf,.ai,.jpg,.png}
\setkeys{Gin}{pagebox=artbox}

\graphicspath{{./acl22-moral-debater-figures/}}

\newcommand{\bsfigure}[3][]{%
	\begin{figure}[t]
		\centering
		\includegraphics[#1]{#2}
		\caption{#3}\label{#2}%
	\end{figure}
}

\RequirePackage{type1cm}
\RequirePackage{color}
\RequirePackage{soul}
\setstcolor{blue}
\definecolor{violet}{rgb}{0.5,0.0,0.5}
\newsavebox\bscombox
\newcommand{\bscom}[3][]{%
	\sbox{\bscombox}{\fontsize{8}{9}\selectfont#1#2#3}
	\noindent
	\st{#2}{\selectfont
		\color{blue}#3\ifx\\#1\\\else{\fontsize{8}{9}\selectfont\color{violet}[#1]}\fi
	}
}

\begin{document}

% If the title and author information does not fit in the area allocated, uncomment the following
%
%\setlength\titlebox{<dim>}
%
% and set <dim> to something 5cm or larger.

\title{The Moral Debater:\\
	 A Study on the Computational Generation of Morally Framed Arguments}

\newcommand{\dlr}{\textsuperscript{$\ddagger$}}
\newcommand{\pb}{\textsuperscript{$\dagger$}}

\author{%
	Milad Alshomary \pb
	\qquad Roxanne El Baff \dlr
	\qquad Timon Gurcke \pb
	\qquad Henning Wachsmuth \pb \\[1.5ex] 
	\pb Paderborn University, Paderborn, Germany,   {\tt milad.alshomary@upb.de}\\
	\dlr German Aerospace Center, Oberpfaffenhofen, Germany,  {\tt Roxanne.ElBaff@dlr.de} \\
}

\maketitle

\begin{abstract}

An audience's prior beliefs and morals are strong indicators of how likely they will be affected by a given argument. Utilizing such knowledge can help focus on shared values to bring disagreeing parties towards agreement. In argumentation technology, however, this is barely exploited so far. This paper studies the feasibility of automatically generating morally framed arguments as well as their effect on different audiences. Following the moral foundation theory, we propose a system that effectively generates arguments focusing on different morals. In an in-depth user study, we ask liberals and conservatives to evaluate the impact of these arguments. Our results suggest that, particularly when prior beliefs are challenged, an audience becomes more affected by morally framed arguments.

\end{abstract}

\section{Introduction}
\label{sec:introduction}

In the last years, more research has been dedicated to studying prior beliefs in argumentation. Understanding the role of prior beliefs helps people craft more effective arguments when targeting a particular audience. Accordingly, operationalizing knowledge about the audience as part of supporting tools for humans or realized in fully automated debating technologies \cite{slonim:2021} could benefit the production of arguments that bridge the gap between disputed parties by focusing on the shared beliefs rather than divisive ones \cite{alshomary:2021b}.

In social psychology, a body of research employs the notion of morals to understand people's judgments on controversial topics~\cite{haidt:2012,fulgoni:2016}. \newcite{feinberg:2015} demonstrated that arguments become more effective when they match the morals of the target audience. Multiple works in computational linguistics have analyzed the persuasive effectiveness of arguments depending on the target audience \cite{durmus:2018, elbaff:2020}, showing that audience-based features reliably predict effectiveness. Different proxies of beliefs have been proposed as part of this, ranging from interests and personality traits~\cite{alkhatib:2020} to stances on popular issues~\cite{alshomary:2021a}. The authors of the latter used the stances to control the generation of argumentative texts. However, they did not assess the effectiveness of these texts on the audience, leaving the importance of encoding beliefs ultimately unclear. Beyond that, little research has been done on generating arguments tailored towards a specific audience, let alone on the importance of morals in achieving agreement.

This work studies the feasibility of generating morally framed arguments computationally and the effect of these arguments on different audiences. For this, we rely on the moral foundation theory \cite{haidt:2012} that projects the moral system into five foundations: care, fairness, loyalty, authority, and purity. To produce arguments that address specific morals, we extend the capabilities of \textit{Project Debater} \cite{slonim:2021}.%
\footnote{Available via an API under: \url{https://early-access-program.debater.res.ibm.com/}}
Project Debater includes a hybrid approach of multiple components designed to generate arguments of high quality that compete with human arguments. Building on this technology helps us focus on evaluating the impact of morally framed arguments, as it ensures a certain base quality level in the generation.

In particular, our proposed extended system takes as input a controversial topic, a stance, and a set of morals. It retrieves a set of argumentative texts, filters the ones conveying the input morals, and finally phrases an argument holding the given stance on the given topic focusing on the given morals. To identify morals in texts, we rely on distant supervision: We use the Reddit dataset of \newcite{schiller:2020}, which contains argumentative texts with annotated aspects, along with the moral-to-concept lexicon of \newcite{hulpus:2020} for the automatic mapping from aspects to morals. Then, we train a BERT-based classifier, achieving high performance on the moral dataset of \newcite{kobbe:2020} compared to ablation baselines.

To assess the effect of morally framed arguments on a particular audience, we consider liberals and conservatives as alternative audiences. We study the question of whether morally loaded arguments are more effective on a specific audience. Additionally, we investigate whether differently-framed moral arguments (targeting different morals) vary in their effect on liberals and conservatives. In line with  \newcite{elbaff:2018}, we design a user study to answer both questions. Our study separately asks three liberals and three conservatives to rank different arguments on specific controversial issues based on how effective they were in challenging or empowering the audience's stance. 

The results suggest that, when arguments challenge the stance, the morally framed ones are generally more effective, especially for the conservative audience. We find that liberals value arguments that focus on their own morals (care and fairness) the most, especially when their stance is challenged. Although conservatives also value respective arguments, a focus on typically conservative morals (loyalty, authority, and purity) becomes more relevant to them when their stance is challenged. Despite the limited size of our user study, these findings hint at the importance of utilizing morals to craft more effective arguments. The code, the trained model, and the data are made publicly available.%
\footnote{\url{https://github.com/webis-de/ACL-22}} The contributions of this work can be summarized as follows: 
\begin{itemize}
	\item A state-of-the-art model for mining statements with defined morals from argumentative texts.
	
	\item A system, based on Project Debater, for generating morally-framed arguments.
	
	\item A user-study giving empirical evidence for the impact of morally-framed arguments on audiences of different political ideologies.
	
\end{itemize}
\section{Related Work}
\label{sec:relatedwork}

\paragraph{Prior Beliefs in Argumentation} 

Studying persuasiveness in argumentation is an active field of research. Besides planning argument content, structure, and style \cite{wachsmuth:2018b}, researchers also considered audience-based features to model persuasiveness. Among these, both \newcite{durmus:2018} and \newcite{elbaff:2020} study how user factors such as religion and political background affect persuasiveness. \newcite{alkhatib:2020} demonstrated that user-based features reflecting beliefs, characteristics, and personality could increase the predictability of argument persuasiveness. Moreover, \newcite{Lukin:2017} demonstrated that persuasiveness correlates with users' personality traits.

The moral foundation theory offers another conceptual model of understanding human judgments in daily life \cite{haidt:2004}. According to the theory, humans subconsciously adhere to five basic moral foundations when judging controversial issues: {\em fairness} (importance of justice, rights, and equality), {\em care} (being kind, and avoiding harm), {\em loyalty} (self-sacrifice, solidarity, belongingness), {\em authority} (respect to traditions and hierarchy), and {\em purity} (sacredness of religion and human). Based on this theory, the disagreement between liberals and conservatives can be explained by the moral gap between the two parties. While liberals rely mainly on care and fairness (so-called {\em individualizing morals}) in their assessment of controversial issues, conservatives consider all moral foundations more evenly, somewhat skewed towards the {\em binding morals} though, that is, loyalty, authority, and purity \cite{graham:2009}. 

Several studies provided evidence of the robustness of the moral foundation theory in understanding people's behaviors and decisions \cite{feinberg:2015,fulgoni:2016, johnson:2018}. For computational purposes, \newcite{kobbe:2020} annotated moral foundations in arguments and analyzed the correlation between morals and argument quality. We use their corpus in our experiments to evaluate our moral classifier.
\paragraph{Argument Generation} 

%\begin{table}[t!]%
%	\centering%
%	\small
%	\renewcommand{\arraystretch}{1}
%	\setlength{\tabcolsep}{1pt}%
%	\begin{tabular}{lccccc}
%		\toprule
%		\bf Topic & \bf Authority & \bf Care  & \bf Fairness & \bf Loyalty & \bf Purity \\
%		\midrule
%		Marijuana Legal. 	  	  & 54\% & 14\% & 13\% & 9\% & 10\%	\\
%		Gun Control 			  & 25\% & 31\% & 26\% & 13\% & 5\%	\\
%		Abortion				  & 21\% & 19\% & 28\% & 14\% & 17\%	\\
%		Death Penalty			  & 7\% & 13\% & 22\% & 21\% & 36\%	\\
%		Minimum Wage			  & 8\% & 16\% & 23\% & 34\% & 19\%	\\
%		Nuclear Energy			  & 2\% & 32\% & 9\% & 20\% & 37\%	\\
%		Cloning					  & 25\% & 20\% & 13\% & 24\% & 17\%	\\
%		School Uniforms			  & 8\% & 10\% & 16\% & 38\% & 28\%	\\
%		\bottomrule
%	\end{tabular}%
%	\caption{Distribution of Moral foundation across topics in the constructed Reddit dataset.}
%	\label{data-stats-table}%
%\end{table}

\begin{table*}[t!]%
	\centering%
	\small
	\renewcommand{\arraystretch}{1}
	\begin{tabular}{lrrrrrrrr}
		\toprule
		\bf	 & \bf Marijuana & \bf Gun & \bf  & \bf Death & \bf Minimum & \bf Nuclear & \bf  & \bf School \\[-0.25ex]
		\bf Moral & \bf Legalization & \bf Control & \bf Abortion & \bf Penalty & \bf Wage & \bf Energy & \bf Cloning & \bf  Uniforms \\
		\midrule
		Care 	 &	14\% & 31\% & 19\% & 13\% & 16\% & 32\% & 20\% & 10\% \\
		Fairness &	13\% & 26\% & 28\% & 22\% & 23\% & 9\% & 13\% & 16\% \\
		Loyalty	 &	9\% & 13\% & 14\% & 21\% & 34\% & 20\% & 24\% & 38\% \\
		Authority&	54\% & 25\% & 21\% & 7\% & 8\% & 2\% & 25\% & 8\% \\
		Purity	 &	10\% & 5\% & 17\% & 36\% & 19\% & 37\% & 17\% & 28\% \\
		\bottomrule
	\end{tabular}%
	\caption{Distribution of the five moral foundations across the eight topics in the constructed Reddit dataset. The topics {\em Cloning} and {\em School Uniforms} are used for validation, all others for training.}
	\label{data-stats-table}%
\end{table*}

Argument generation approaches have been proposed for a spectrum of different tasks. \newcite{hua:2019} and \newcite{alshomary:2021c} worked on counter-argument generation, tackling the task by rebutting an argument's conclusion or by undermining one of its weak premises, respectively. Others opposed a given claim \cite{bilu:2015, hidey:2019}, generated argumentative claims on a given topic controlled for certain aspects \cite{schiller:2020}, or reconstructed implicit conclusions \cite{alshomary:2020a, syed:2021}. However, none of these approaches considered generating augmentative texts that target a specific audience. 

Recently, \newcite{alshomary:2021a} introduced the task of belief-based claim generation, aiming to generate claims on a given topic that match a given target audience. As a model of beliefs, they considered people's stances on big issues. While they demonstrated the feasibility of encoding beliefs into claims, they did not study the effectiveness of these claims on an audience. Another driver in the field of argument generation is {\em Project Debater} by \newcite{slonim:2021}. Project Debater is an end-to-end system for argument mining, retrieval, and generation. The system relies on a hybrid approach consisting of retrieval, mining, clustering, and rephrasing components. In their manual evaluation, \newcite{slonim:2021} observe a competent quality of generated arguments compared to those crafted by humans. To generate {\em morally framed} arguments, we extend their system. Relying on Project Debater helps us alleviate confounding effects of argument quality and, so, to focus on testing whether moral utilization affects the audience.
\section{Moral Foundation Classification}
\label{sec:classifier}

Our proposed system relies on the ability to identify morals in arguments automatically. Existing approaches to mining morals from texts are either lexicon-based or machine learning-based. A number of datasets with morals have been constructed for domains such as social media or news articles. For argumentative texts,  \newcite{kobbe:2020} manually annotated a small dataset of 220 arguments, which is only suitable for evaluation. We, therefore, decided to develop a moral foundation classifier based on data collected automatically using distant supervision, as explained in the following.

\subsection{Data Collection} 

To circumvent the need for annotated data, we construct a training dataset following a distant-supervision approach. In particular, moral foundations are revealed as aspects of concerns in discussions of controversial topics. For example, when discussing \textit{School Uniform} from the \textit{authority} perspective, aspects such as \textit{respect} and  \textit{obedience} often arise. Given this observation, we referred to the dataset of \newcite{schiller:2020} which contains short argumentative texts on eight topics along with aspects annotated automatically for each text. We then assigned each text a set of moral foundations based on the aspects appearing in the text. To map aspects to moral foundations, we employed the lexicon of \newcite{hulpus:2020} which connects moral foundations to Wikipedia concepts. After filtering out arguments without any mapping and balancing the data across the five moral foundations, this resulted in a dataset with 230k argumentative texts and the corresponding morals. We split the dataset into six topics for training and two for validation (testing will happen on other data below). Details on the distribution of the morals across topics are found in Table~\ref{data-stats-table}. 

To assess the quality of the distantly supervised dataset, two authors of the paper manually evaluated the correctness of the assigned morals on a sample of 100 examples. 77\% of the cases were considered correct by at least one author, 44\% by both. The Cohen's $\kappa$ agreement was 0.32, which is not high, but in line with other subjective argument-related annotations \cite{elbaff:2018}. Table~\ref{data-argument-examples-table} shows example sentences with assigned morals from the dataset.

\begin{table}[t!]%
	\centering%
	\small
	\renewcommand{\arraystretch}{1}
	\setlength{\tabcolsep}{3pt}%
	\begin{tabular}{p{6cm}l}
		\toprule
		\bf Argumentative Sentence & \bf Moral\\
		\midrule
		Abortion isn't murder because abortion is legal and murder is an illegal killing of another person.
		 & Care\\
		\addlinespace
		The simple fact, is if your wages are not fair, you have the right to unionize and get better wages. & Fairness \\
		\addlinespace
	This intertribal coalition also pushed for a new standard for national monuments and tribal involvement. & Loyalty \\	
% 	    Americans. This intertribal coalition also pushed for a new standard for national monuments and	& Loyalty \\	
 	   \addlinespace
 	    Gun laws are only obeyed by law abiding people. &  Authority \\
 		\addlinespace
		People that ignore the social contract and human decency aren't deterred by any punishment. & Purity\\
		\bottomrule
	\end{tabular}%
	\caption{Five example argumentative sentences from our dataset with the morals assigned automatically.}
	\label{data-argument-examples-table}%
\end{table}

\subsection{Approach} 

We rely on a BERT-based classifier to identify morals in texts \cite{devlin:2019}, starting from the pre-trained \textit{bert-based-cased} model. We fine-tuned the model on our training set for three epochs with a batch size of 16 and a learning rate of $3 \cdot e^{-5}$. In the training phase, the input was an argumentative sentence and the corresponding moral foundation. Since an argument may contain multiple sentences, each reflecting a specific moral, an argument's final set of morals consists of all sentences' morals predicted with confidence above 0.5. To assess the classifier's effectiveness more reliably, we trained six models on different random samples of size 50k and computed their average F$_1$-score.

\subsection{Experimental Setup} 

\begin{table*}[t!]%
	\centering%
	\small
	\renewcommand{\arraystretch}{1}
	\setlength{\tabcolsep}{3.25pt}%
	\begin{tabular}{lrrrrrrrrrrrrrrrcrrr}
		\toprule
		 & \multicolumn{3}{c}{\bf Care} & \multicolumn{3}{c}{\bf Fairness}  & \multicolumn{3}{c}{\bf Loyalty} & \multicolumn{3}{c}{\bf Authority} & \multicolumn{3}{c}{\bf Purity} && \multicolumn{3}{c}{\bf Macro} \\
		\cmidrule(l{5pt}r{5pt}){2-4} 	
		\cmidrule(l{5pt}r{5pt}){5-7}
		\cmidrule(l{5pt}r{5pt}){8-10}
		\cmidrule(l{5pt}r{5pt}){11-13}
		\cmidrule(l{5pt}r{5pt}){14-16}
		\cmidrule(l{5pt}r{5pt}){18-20}
		\bf  Approach & \bf Pre & \bf Rec & \bf F$_1$ & \bf Pre & \bf Rec & \bf F$_1$ & \bf Pre & \bf Rec & \bf F$_1$ & \bf Pre & \bf Rec & \bf F$_1$	& \bf Pre & \bf Rec & \bf F$_1$ && \bf Pre & \bf Rec & \bf F$_1$ \\
		\midrule
		Lexicon & 0.64     & \bf 0.88 & \bf 0.60 & 0.07     & \bf 0.70  & 0.13    & 0.09    & \bf 0.86 & 0.17     & 0.14     & 0.63     & 0.23     & 0.16     & \bf 0.72 & 0.27 && 0.18 & \bf 0.76 & 0.28  \\
		mBERT   & \bf 0.74 &     0.38 & 0.50    & \bf 0.47 &     0.35 & \bf 0.40 & \bf 0.50 & 0.10      & 0.16     & \bf 0.43 & 0.09     & 0.14     & \bf 0.56 & 0.13     & 0.21 && \bf 0.54 & 0.21 & 0.28 \\
		\addlinespace
		Ours    & 0.54     &     0.56 & 0.52    & 0.31     &     0.55 & 0.37    & 0.21    & 0.54     & \bf 0.28 & 0.23     & \bf 0.74 & \bf 0.34 & 0.46     & 0.48     & \bf 0.46 && 0.35 & 0.57 & \bf 0.40 \\
		\bottomrule
	\end{tabular}%
	\caption{Moral foundation classification: Precision (Pre), recall (Rec), and F$_1$-score (F$_1$) of our approach and the baselines for each moral foundation as well as the macro averages. The best value in each column is marked bold.}
	\label{moral-classifier-eval}%
\end{table*}

%\begin{table}[t!]%
%	\centering%
%	\small
%	\renewcommand{\arraystretch}{1}
%	\setlength{\tabcolsep}{2.5pt}%
%	\begin{tabular}{lrrrrr@{\quad}r}
%		\toprule
%		\bf Approach          	& \bf Care & \bf Fairn.  & \bf Loyalty & \bf Auth. & \bf Purity & \bf Macro \\
%		\midrule
%		Lexicon 	  		 	&  \bf 0.60 &  0.13  &  0.17	& 0.27  & 0.23 & 0.28 \\
%		mBERT        			&  0.50 &  0.4  & 0.16   & 0.14	& 0.21 & 0.28 \\
%		\addlinespace
%		Ours                  	&  0.53 & \bf 0.37  & \bf 0.28 & \bf 0.34	& \bf 0.46 & \bf 0.40 \\ 
%		\bottomrule
%	\end{tabular}%
%	\caption{F$_1$-score of our approach and the baselines for each moral foundation as well as the macro F$_1$-score.}
%	\label{moral-classifier-eval}%
%\end{table}

For comparison, we consider two baselines. The first is the model performing best in the experiments of \newcite{kobbe:2020}, which is a multi-label BERT-based model trained on the Twitter moral corpus of \newcite{hoover:2020}. We trained our own version on the same dataset and referred to it as \textit{mBERT}. The second baseline is a simple lexicon-based approach that computes the frequency of words belonging to each of the moral foundations \cite{araque:2020}, called \textit{Lexicon} below. As a lexicon, we used the \texttt{moralstrength} library.%
\footnote{Link: \url{https://github.com/oaraque/moral-foundations}}. 
%The predicted moral then would be the one with the most occurring relevant words.

\begin{table*}[t!]%
	\centering%
	\small
	\renewcommand{\arraystretch}{1}
	\setlength{\tabcolsep}{6pt}%
	\begin{tabular}{p{5.5cm}lp{3cm}ll}
		\toprule
		\bf Argument & \bf Ground truth & \bf Lexicon & \bf mBERT & \bf Our approach \\
		\midrule
		This is just wrong we should not insult who we believe in we do not need to know what you people think.
		& Authority & Care, fairness, authority, purity & -- & Authority \\
		\addlinespace
		Christianity does offer hope in the world. Christianity does tell others to help the poor. & Care & Care, fairness, loyalty, authority, purity &	Care & Authority, loyalty \\
		\bottomrule
	\end{tabular}%
	\caption{Ground-truth moral foundations of two example arguments from the dataset of \newcite{kobbe:2020} in comparison to the morals assigned by the two classification baselines (Lexicon, mBERT) and by our approach.}.
	\label{example-assigned-morals}%
\end{table*}

We tested all models on the dataset of \newcite{kobbe:2020}, which consists of 220 arguments annotated for moral foundations by two annotators.

\subsection{Results} 

Table \ref{moral-classifier-eval} shows the F$_1$-score of all evaluated models for each moral as well as the macro F$_1$-score. Additionally, we show the precision and recall for each approach. In terms of F$1$-score, our approach outperforms both baselines across three of the five moral foundations as well as on average. We observe that effectiveness varies in terms of precision and recall between the Lexicon and the mBERT baseline. The stable effectiveness of our approach across the five morals signals the advantage of the proposed dataset that we used in our approach. Hence, we use this model later for morally framed argument generation in our system. 

Table \ref{example-assigned-morals} shows two example arguments with the manually annotated morals along with the ones predicted by the baselines and by our approach. We see that the Lexicon baseline assigns all morals to each argument most of the time, leading to the high recall across all morals. The first row of the table shows an example argument from the test set in which our approach was able to detect its \textit{authority} moral while mBERT failed. In the second row, our approach missed the \textit{care} moral in the argument but highlighted \textit{loyalty}, a moral that probably emerges from the aspect of helping each other.
\section{The Moral Debater System}
\label{sec:system}

\bsfigure{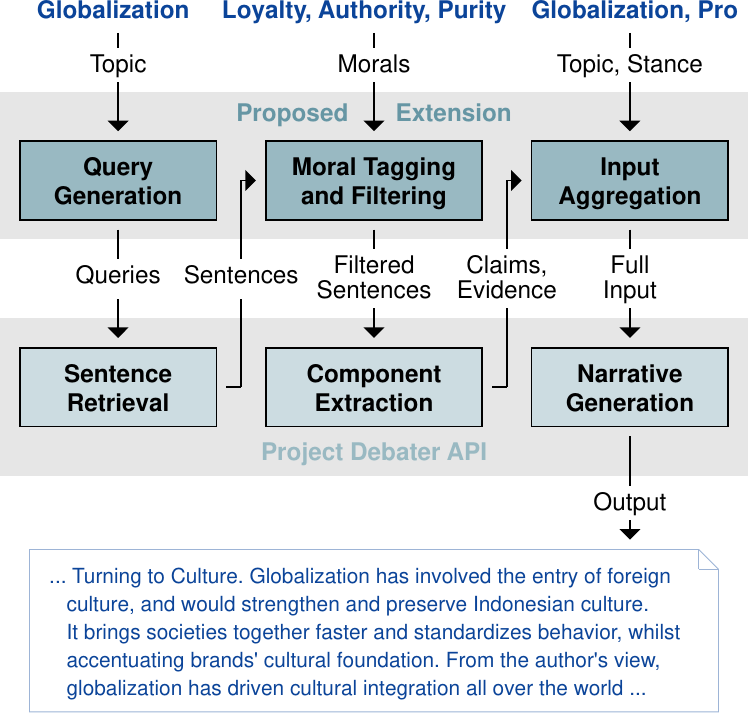}{High-level process of the proposed moral debater system, which extends the capabilities of Project Debater by moral tagging and filtering, in order to output a morally framed argument for a given topic, a stance on the topic, and a set of morals.}

This section describes the system that we developed to study the effect of morally framed arguments on the audience. Our design extends the capabilities available via the Project Debater API with the moral foundation classifier from Section~\ref{sec:classifier}. As input, it takes a controversial topic (say, ``globalization''), a stance on the topic (say, {\em pro}), and a set of morals to be targeted (say, {\em loyalty}, {\em authority}, and {\em purity}). Then, it determines a collection of claims and evidence on the given topic that focus on the given morals, from which it constructs an argument with the given stance. Figure~\ref{approach.ai} shows the high-level process of the proposed system. We describe the three-step process in detail in the following, highlighting our proposed integration.

\subsection{Query Generation \& Sentence Retrieval} 
First, the system retrieves a collection of argumentative sentences discussing the controversial topic from Project Debater's index, which contains 400 million news articles. The articles are split into sentences and indexed along with several meta-annotations. We generate several queries containing only the topic keywords without any topic expansion to focus on relevant sentences. We restrict the retrieved sentences to only those annotated as having sentiment or causality markers. Section 5 gives more details on the constructed queries.

\subsection{Moral Tagging and Filtering \& Component Extraction} 

Second, the trained classifier is used to annotate each argumentative sentence for all likely moral foundations. It then filters out those sentences that either does not have any moral or contain at least one moral not given as input. Next, through Project Debater's API, the system generates for each of the remaining sentences a likelihood score reflecting whether it contains a claim or evidence following the approach of \newcite{ein:2020}. We instruct the API to keep only sentences having a claim with likelihood higher than \texttt{claim\_threshold} or evidence with likelihood higher than \texttt{evidence\_threshold} (the exact thresholds are given in Section~\ref{sec:evaluation}). Additionally, the API identifies claim boundaries for sentences containing claims and extracts the exact span of text containing the claim.

\begin{table}[t!]%
	\centering%
	\small
	\renewcommand{\arraystretch}{1.0}
	\setlength{\tabcolsep}{2.5pt}%	
	\begin{tabular}{p{0.98\columnwidth}}
		\toprule
		{\bf Topic: Globalization}  \\
		\midrule
		{\bf Binding argument:} The crowd raised four issues, explaining its views. The first claim is that globalization is reducing the importance of nation-states. The next issue will show how Globalization and structural forces aggravate poverty. In addition, we will hear about pollution and Culture. 
		
		$\cdots$
		
		Lastly, Culture.  Globalization has destabilized previously immutable social institutions, shifting cultural value away from old traditions to new more individualistic and market friendly ideas.   It is often said to have a negative effect on the world's cultural diversity.   Cultural and geographical dimensions of transformational leadership become blurred as globalization renders ethnically specific collectivist and individualistic effects of organizational behavior obsolete in a more diversified workplace.  \\
		\midrule
		{\bf Individualizing argument:} 
		The crowd raised four issues, explaining its views. The first claim is that Globalization on its own cannot end gender inequality. In addition, we will hear about harm, economy and processes. 
		
		Starting with gender inequality.  There are various studies available that depict globalization as a hindrance toward gender inequality.   Globalization on its own cannot end gender inequality.  
		
		Turning to harm. $\cdots$  Globalization is a threat to culture and religion, and it harms indigenous people groups while multinational corporations profit from it. It has been criticized for benefiting those who are already large and in power at the risk and growing vulnerability of the countries' indigenous population.
		$\cdots$
		 \\
		\midrule
		{\bf Uncontrolled argument:} 
		The crowd raised four issues, explaining its views. The first claim is that globalisation creates economic and cultural imbalances in developing nations. The next issue will show how globalization is reducing the importance of nation-states. And the third point is that globalization is a threat. In addition, we will hear about processes. \\
		
		\\
		Starting with economy.  Globalization does not work for all the economies that it affects, and that it does not always deliver the economic growth that is expected of it.   Globalisation and neoliberalism have exacerbated already unequal economic relations.   Although globalization takes similar steps in most countries, scholars such as Hodge claim that it might not be effective to certain countries and that globalization has actually moved some countries backward instead of developing them.  
		
		$\cdots$
		 \\
		\bottomrule
	\end{tabular} 
	\caption{
		Example generated arguments against \textit{Globalization} for different focused morals. The '$\cdots$' indicates an omitted content due to space limitation. Full arguments are shown in the appendix.
	}
	\label{table-example-generated-counters}
\end{table}

\subsection{Input Aggregation \& Narrative Generation} 

Third, our proposed extension aggregates the given list of claims and evidence sentences with the input topic and stance. It then uses Project Debater's narrative generation API to generate the final argument. The narrative generation identifies the stance of claims and evidence towards the topic according to the approach of \newcite{bar:2017}. Only those matching the input stance are kept. Redundant elements are then filtered out, and the remaining ones are grouped into thematic clusters, where a theme is a Wikipedia title \cite{slonim:2005}. The process of building these clusters also includes extracting one claim that represents the theme. Each theme will then be represented by a paragraph in the output argument. Finally, a set of algorithms is used to perform various kinds of re-phrasing on the argument level (e.g., pronoun resolution) and on the paragraph level (e.g., ensuring that different arguments are put together)~\cite{slonim:2021}. 

Example arguments with and without controlled morals are shown in Table~\ref{table-example-generated-counters}. By concept, each argument starts with an introductory paragraph listing the main themes of discussion, followed by a set of paragraphs, each combining claims and evidence on one theme.
\section{Studying the Effect of Moral Framing}
\label{sec:evaluation}

To evaluate our hypothesis on the effect of morally framed arguments, we carried out a user study with two opposing target audiences, {\em liberals} and {\em conservatives}. Our primary goal was to investigate whether morally framed arguments are more effective than uncontrolled ones. Additionally, we sought to determine whether differently-framed arguments affect liberals and conservatives differently. In the following, we report on this study.

\subsection{Experimental Setup}

\paragraph{Arguments} 

We considered ten popular topics from the website debate.org, called {\em big issues} there. For each topic, we used our system to construct three arguments: one argument focusing on care and fairness (\textit{individualizing}), one focusing on loyalty, authority, and purity (\textit{binding}), and one baseline argument where we did not control the morals targeted (\textit{uncontrolled}). We created arguments separately for both stances (pro and con), resulting in a total of $10 \cdot 3 \cdot 2 = 60$ arguments.

To construct each argument, we used the following parameters. For each topic, we built four queries, retrieving 10k sentences with 6 to 60 tokens per query. The first query retrieved sentences containing the topic. The second and third query targeted claim-like sentences, requiring the occurrence of (a)~at least one causality marker or (b)~both causality and a sentiment marker. Each needed to appear together with the topic in a window of 12 tokens. The last query aimed to retrieve evidence by filtering only those sentences that contained any of the following tokens: ``surveys'', ``analyses'', ``researches'', ``reports'', ``research'', and ``survey''. A moral was assigned to a retrieved sentence if the probability of our classifier was higher than 0.5. After initial tests, we set the \texttt{claim\_threshold} and \texttt{evidence\_threshold} to 0.8 and 0.6 respectively. We left all other settings to the default values of Project Debater's API.

\paragraph{Internal Study on Argument Quality}

Before we launched our main study, two authors of this paper manually assessed the quality of the generated arguments and the morals addressed in each. In particular, each of them read all 60 arguments and ranked their {\em relevance}, {\em coherence}, and {\em argumentativeness} on a 5-point Likert scale. While reading each argument, they also highlighted spans of text that they found to reflect a specific moral.

\paragraph{External Study on Argument Effectiveness} 

To answer our research questions, we conducted a two-phase user study on the platform {\em Upwork}: First, we determined the political ideology of each participant, and then, we let selected participants rank the different arguments.

In the first phase, we asked people living in the US that are experienced in writing and content editing to perform the {\em Political Typology Quiz}, available through the Pew Research Center, in order to identify their political ideology.%
\footnote{Political Typology Quiz: \url{ https://www.pewresearch.org/politics/quiz/political-typology/}} 
In 17 questions, the quiz asks participants to state their views on controversial issues in the US. The test results place the participants on a spectrum of ideologies from {\em solid liberal} (left) to {\em core conservative} (right). 

In the second phase, we chose only six participants from the first phase due to budget constraints, three solid liberals (one male, two female) and three core conservatives (two males, one female). We showed each of them three arguments (one individualizing, one binding, one uncontrolled) for all 20 topic-stance pairs. For each pair, the participants read the three arguments and ranked them by perceived {\em effectiveness}. We followed \newcite{elbaff:2018}, defining the effectiveness of an argument either by how empowering it is (if the participant has the same stance on the topic) or by how challenging it is (otherwise). For this purpose, the participants self-assessed their stances on each topic on a 5-point Likert scale, from 1~(strongly~disagree) to 5~(strongly support) before reading the arguments.\footnote{Given an estimated workload of~3 to 3.5~hours, we paid each participant a fixed rate of \$75.}

\subsection{Results}

In the following, we present the results of both studies, attempting to answer our research questions.

\begin{table}[t!]%
	\centering%
	\small
	\renewcommand{\arraystretch}{1}
	\setlength{\tabcolsep}{3pt}%
	\begin{tabular}{l@{\hspace*{-1.6cm}}rrrrrrrr}
		\toprule
		\bf Type & \bf Argumentativeness & \bf Relevance  & \bf Coherence  \\
		\midrule
		Binding arguments			& 3.8 & 3.8 & \bf 4.0   \\
		Individualizing arguments 		& \bf 4.2 & 4.0 & 3.9   \\
		Uncontrolled arguments		& 4.1 & \bf 4.1 & 3.9  \\
		\bottomrule
	\end{tabular}%
	\caption{Mean quality scores of the three types of evaluated arguments on a 5-point scale (higher is better).}
	\label{internal-argument-manual-evaluation-table}%
\end{table}

\begin{table}[t!]%
	\centering%
	\small
	\renewcommand{\arraystretch}{1}
	\setlength{\tabcolsep}{2pt}%
	\begin{tabular}{lrrrrrrrr}
		\toprule
		\bf Type  & \bf Care & \bf Fairness & \bf Loyalty & \bf Authority & \bf Purity \\
		\midrule
		Binding 	& 16\% & 17\% & 10\% & 47\% & 9\% \\
		Individualizing  	& 17\% & 36\% & 6\% & 35\% & 6\% \\
		Uncontrolled	&	11\% & 21\% & 4\% & 54\% & 10\% \\
		\bottomrule
	\end{tabular}%
	\caption{Distribution of the five moral foundations found in the three types of evaluated arguments.}
	\label{internal-argument-manual-evaluation-table-2}%
\end{table}

% OLD.
%%%\begin{table}[t!]%
%%%	\centering%
%%%	\small
%%%	\renewcommand{\arraystretch}{1}
%%%	\setlength{\tabcolsep}{1.55pt}%
%%%	\begin{tabular}{lrrrrrrrr}
%%%		\toprule
%%%		\bf Type & \bf Rel.  & \bf Coh. & \bf Arg.  & \bf Care & \bf Fairn. & \bf Loyal. & \bf Auth. & \bf Pur. \\
%%%		\midrule
%%%		Binding 			& 3.8 & 4.0 & 3.8 & 16\% & 17\% & 10\% & 47\% & 9\% \\
%%%		Individ. 			& 4.0 & 3.9 & 4.2 & 17\% & 36\% & 6\% & 35\% & 6\% \\
%%%		Uncont.		 		& 4.1 & 3.9 & 4.1 & 11\% & 21\% & 4\% & 54\% & 10\% \\
%%%		\bottomrule
%%%	\end{tabular}%
%%%	\caption{Mean quality assessment scores (on a 5-point Likert scale) of the 60 evaluated arguments, as well as the moral-foundation distribution (the last five columns).}
%%%	\label{internal-argument-manual-evaluation-table}%
%%%\end{table}

\paragraph{Argument Quality} 

Table~\ref{internal-argument-manual-evaluation-table} presents the quality scores  for each argument type and Table~\ref{internal-argument-manual-evaluation-table-2} the distribution of moral foundations. Comparing the scores of binding and individualizing arguments to the uncontrolled ones, we see that our method did not notably worsen the quality of the generated arguments. The moral foundation distribution indicates that binding arguments have a relatively higher focus on loyalty, authority, and purity than the individualizing arguments and a lower focus on fairness and care. This supports the impact of our method on controlling morals in arguments.

\bsfigure{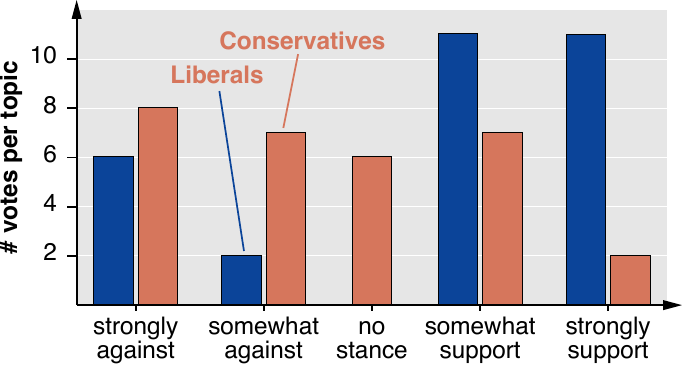}{Distribution of the stances of liberals and conservatives on the ten given topics, on a 5-point Likert scale from 1 (strongly against) to 5 (strongly support).}

\paragraph{Empowering vs. Challenging} 

Figure~\ref{stance-distribution} shows the distribution of challenging and empowering arguments. Liberals were more decisive with their stance on the given topics, with 73\% being on the pro side, whereas only 30\% of the conservatives were on that side (50\% con side, 20\% no stance). Since we presented arguments for both sides for each topic, we had an equal distribution of empowering and challenging arguments for the liberals. However, for conservatives, we had 40\% empowering and 40\% challenging arguments due to the 20\% undecided cases. Since arguments supporting one side of a debate are rather challenging for the undecided audience, in our analysis below, we consider the 20\% undecided cases to be challenging.

%\begin{table}[t!]%
%	\centering%
%	\small
%	\renewcommand{\arraystretch}{1}
%	\setlength{\tabcolsep}{3pt}%
%	\begin{tabular}{lrrrr}
%		\toprule
%		\bf Moral Frame      & \bf Rank 1 & \bf Rank 2  & \bf Rank 3 & \bf AVG. Rank \\
%		\midrule
%		Binding 	  		&  24\% &  32\%  &  44\% & 2.20  \\
%	    Individualizing     &  45\% &  38\%  &  17\% & 1.72 \\
%		Uncontrolled        &  31\% &  30\%  &  39\% & 2.08 \\
%		\bottomrule
%	\end{tabular}%
%	\caption{Rank distribution of all the six annotators (liberals and conservatives) as well as the average rank, for each type of arguments (binding, individualizing, and uncontrolled)}
%	\label{manual-main-eval-table}%
%\end{table}

\begin{table}[t!]%
	\centering%
	\small
	\renewcommand{\arraystretch}{1}
	\setlength{\tabcolsep}{3pt}%
	\begin{tabular}{ll@{$\!\!\!$}rrrr}
		\toprule
		\bf Ideology & \bf Morals  & \bf Rank 1 & \bf Rank 2  & \bf Rank 3 & \bf Mean \\
		\midrule
		 Liberals & Binding &  23\% &  37\%  &  40\% & 2.17  \\
		 & Individualizing     		&  40\% &  40\%  &  20\% & \textsuperscript{*}\bf 1.80 \\
		 & Uncontrolled      		&  37\% &  23\%  &  40\% & 2.03 \\
		 \addlinespace
		 Conser- & Binding &  25\% &  27\%  &  48\% & 2.23  \\
		 vatives & Individualizing     		&  50\% &  37\%  &  13\% & \textsuperscript{**}\bf 1.63 \\
		 & Uncontrolled      		&  25\% &  37\%  &  38\% & 2.13 \\	
		 \addlinespace
		 All & Binding 	  	&  24\% &  32\%  &  44\% & 2.20  \\
		 & Individualizing     		&  45\% &  38\%  &  17\% & \bf \textsuperscript{**}1.72 \\
		 & Uncontrolled      		&  31\% &  30\%  &  39\% & 2.08 \\	 
		\bottomrule
	\end{tabular}%
	\caption{Rank distribution and the mean rank for each type of moral framing (binding, individualizing, uncontrolled) reflecting the effectiveness according to the different participant groups (liberals, conservatives, all). Values marked with * and ** are significantly better than Uncontrolled ranks at $p < 0.1$ and $p < 0.05$ respectively.}
	\label{manual-main-eval-table}%
\end{table}

\paragraph{Effectiveness of Moral Arguments} 

Table \ref{manual-main-eval-table} shows the rank distribution for morally-framed arguments (\textit{binding} and \textit{individualizing}) compared to the \textit{uncontrolled} ones for liberals, conservatives, and all together. In general, the participants ranked the arguments framed in terms of fairness and care (\textit{individualizing}) significantly better than the \textit{uncontrolled} ones, with an average rank of 1.72 compared to 2.08.%
\footnote{The difference in mean ranks is 0.36, 95\% CI [0.17, 0.57].} 
This difference is significant at $p < 0.05$ using student $t$-test. This signals a positive answer to our first research question: a focus on morals can make arguments more effective. A closer look at the distribution of arguments at \textit{Rank 1} shows that conservatives were more susceptible to moral arguments (75\% binding and individualizing) compared to liberals (63\%).

Next, we examine whether arguments with different morals affect liberals and conservatives differently by looking at the achieved ranks of both empowering and challenging arguments.

\begin{table}[t!]%
	\centering%
	\small
	\renewcommand{\arraystretch}{1}
	\setlength{\tabcolsep}{3.5pt}%
	\begin{tabular}{ll@{\hspace*{-2em}}rr@{\hspace*{1.5em}}r}
		\toprule
		\bf Ideology & \bf Moral & \bf Empowering  & \bf Challenging & \bf Both\\
		\midrule
		\addlinespace
		Liberals & Binding  		& 2.27 & 2.07 & 2.17 \\
		& Individualizing 			& \bf 1.83 & \textsuperscript{*}\bf 1.77 & \bf \textsuperscript{*}1.80 \\
		&Uncontrolled    			& 1.90 & 2.17 & 2.03 \\
		\addlinespace
		Conser- & Binding  			& 2.29  & 2.19  & 2.23 \\
		vatives & Individualizing 	& \bf 1.71  & \textsuperscript{*}\bf 1.58  & \textsuperscript{**}\bf 1.63 \\
		&Uncontrolled    			& 2.00  &  2.22 & 2.13 \\
		\addlinespace
		All & Binding  				& 2.29  & 2.19  & 2.20 \\
		& Individualizing 			& \bf 1.71  &  \textsuperscript{**}\bf 1.58  & \textsuperscript{**}\bf 1.72 \\
		&Uncontrolled    			& 2.00	&  2.22 & 2.08 \\
		\bottomrule
	\end{tabular}%
	\caption{The mean rank of each type of moral framing (binding, individualizing, uncontrolled) according to the different participants (liberals, conservatives, all) for challenging arguments (opposite stance to participant), empowering arguments (same stance), and both. Values marked with * and ** are significantly better than Uncontrolled ranks at $p < 0.1$ and $p < 0.05$ respectively.}
	\label{manual-main-eval-table-all}%
\end{table}

\paragraph{Effectiveness depending on Ideology}

Looking at the mean ranks assigned by liberals in Table \ref{manual-main-eval-table-all}, we observe that challenging arguments that focus on individualizing morals (care and fairness) are most effective. We validate that the difference is significant for $p < 0.1$ using student $t$-test. This is in line with \newcite{feinberg:2015} who found that arguments framed in terms of liberal morals were more convincing to liberals. Notably, this effectiveness slightly decreases when arguments are empowering. A reasonable hypothesis is that, in the case of empowering arguments, the audience may be more interested in the opposing views, which might be covered by uncontrolled arguments. We investigate this hypothesis further via a follow-up questionnaire below. 

Now, we look at the conservatives. Although they also valued the individual arguments the most, we observe that, when arguments challenged their views, a focus on binding morals (loyalty, authority, and purity) became slightly more effective than the uncontrolled arguments. Generally, morally framed arguments that challenged the views of conservatives were significantly more effective than uncontrolled ones at $p < 0.1$ using the student $t$-test.

\paragraph{Agreement across Ideologies} 

We measured inter-annotator agreement between the participants using Kendall's $W$~\cite{kendall:1939}. The agreement of all six participants was 0.29. In contrast, when considering liberals and conservatives separately, it increased to 0.35 for liberals and 0.51 for conservatives. This indicates higher agreement between participants having similar political ideology and matches the common notion that conservatives are more unified in their views than liberals.

\paragraph{Reasons behind Effectiveness Judgments} 

In a follow-up questionnaire, we investigated our participants' judgments. We asked them to self-assess whether they prefer (1)~arguments with {\em knowledge} they have or are not familiar with, (2)~arguments that matched or challenged their {\em own views}, (3) arguments that convince {\em others} who share or oppose their {\em views}, and (4)  what affected the judgments of argument {\em effectiveness} more: knowledge or views, each in empowering and challenging cases.

Table~\ref{followup-study-analysis-table} shows that the participants ranked knowledge as the most relevant {\em effectiveness} aspect. In terms of {\em others' views}, the majority valued arguments that focus on the opposing views, whereas preferences differ for empowering and challenging arguments on {\em own views}. Due to the reliability issues of self-assessment of one's moral judgments \cite{pizarro:2000}, we acknowledge the limitation of this study, though. We present details on the questionnaire and its results in the appendix.

\begin{table}[t!]%
	\centering%
	\small
	\renewcommand{\arraystretch}{1}
	\setlength{\tabcolsep}{2pt}%
	\begin{tabular}{ll@{\hspace*{-0.75cm}}rrr}
		\toprule
		& & \bf Empowering & \bf Challenging & \bf All \\
				\midrule
		Knowledge   & Know about   & 33.3\%     & 0.0\%      & 16.7\% \\ 
						& Not familiar & \bf 66.7\% & \bf 83.3\% & \bf 75.0\% \\ 
						& Neither 	   & 0.0\%      &  16.7\%    &  8.3\% \\
\midrule
		Own views  & Matched      & \bf 50.0\% & 16.7\%     & 33.3\% \\ 
						& Challenging  & 33.3\%     & \bf 50.0\% & \bf 41.7\% \\ 
						& Neither      & 16.7\% & 33.37\%   & 25.0\% \\		
		\midrule
		Others' views & Share view  &  16.7\% &  0.0\% &  8.3\% \\ 
						& Oppose view &  \bf 66.7\% &  \bf 66.7\% &  \bf 66.7\% \\ 
						& Neither 	  &  16.7\% &  33.3\% &  25.0\% \\
		
		\midrule
		Effectiveness   & Knowledge &  \bf 83.3\% &  \bf 66.7\% &  \bf 75.0\% \\ 
							& Views     &  16.7\% &  33.3\% &  25.0\% \\ 
							& Neither   &  0.0\% &  0.0\% &  0.0\% \\
		\bottomrule
	\end{tabular}%
	\caption{Distribution of preferences (options) selected by the annotators for each of the four asked questions for empowering and challenging cases.}
	\label{followup-study-analysis-table}%
\end{table}

%\begin{table}[t!]%
%	\centering%
%	\small
%	\renewcommand{\arraystretch}{1}
%	\setlength{\tabcolsep}{3pt}%
%	\begin{tabular}{llrr}
%		\toprule
%		& & \bf Empowering & \bf Challenging \\
%		\midrule
%		\bf Own views  & Matched      & \bf 50.0\% & 16.7\% \\ 
%		& Challenging  & 33.3\%     & \bf 50.0\%  \\ 
%		& Neither      & 16.7\% & 33.37\%   \\
%		
%		\midrule
%		\bf Knowledge   & Know about   & 33.3\%     & 0.0\%   \\ 
%		& Not familiar & \bf 66.7\% & \bf 83.3\%  \\ 
%		& Neither 	   & 0.0\%      &  16.7\%    \\
%		
%		\midrule
%		\bf Others' views & Share view  &  16.7\% &  0.0\% \\ 
%		& Oppose view &  \bf 66.7\% &  \bf 66.7\% \\ 
%		& Neither 	  &  16.7\% &  33.3\% \\
%		
%		\midrule
%		\bf Effectiveness   & Knowledge &  \bf 83.3\% &  \bf 66.7\% \\ 
%		& Views     &  16.7\% &  33.3\% \\ 
%		& Neither   &  0.0\% &  0.0\%\\
%		\bottomrule
%	\end{tabular}%
%	\caption{Distribution of preferences (options) selected by the annotators for each of the four asked questions for both the empowering and challenging cases.}
%	\label{followup-study-analysis-table}%
%\end{table}

\section{Discussion and Limitations}

We explicitly acknowledge the limited number of liberals and conservatives that we recruited in our final user study, who might not represent the whole population. Since the low sample size affects the reliability of our results, we performed significance tests to report our main results with a certain confidence level. As far as budget permits, further studies should be run with a more significant sample to reassess the reliability of the results.

Despite the limitation above, the results of our evaluation indicate that arguments targeting the moral foundations of care and fairness are more effective than others, at least in the tested sample. Also, we observed that focusing on the audience's views in challenging arguments makes them more effective. Regarding political ideologies, while our results match the literature in that liberals were affected more by arguments focusing on their views (care and fairness), especially when challenged, conservatives did not rank binding arguments higher than individualizing ones. However, the conservatives still showed a higher tendency to be affected by morally framed arguments than liberals.

Our approach is limited by its capability to retrieve argumentative texts that discuss the topic from different moral perspectives. In the retrieval component, we have focused on obtaining sentences containing precisely the addressed topic. A more elaborated approach could broaden the search to relevant topics through topic expansion. Further, we could use \textit{Project Debater} only through the predefined API, restricting the way we integrate the moral tagging of sentences. Ideally, it would be performed during indexing; then, choosing the morals to focus on could be defined as part of the queries. Also, it is noteworthy that our moral classifier is trained on an automatically annotated dataset which may have limited its performance. 

Still, we demonstrated that it is feasible to tune arguments automatically to target certain morals and that such arguments tend to be more effective. Moreover, our results indicate that differently framed arguments have a different effect on different audiences. This opens opportunities for generating more effective arguments when targeting a specific audience, bridging the gap between disagreeing parties by focusing on the shared beliefs.
\section{Conclusion}

In this work, we have proposed an extension of Project Debater that generates morally framed arguments. Due to the lack of training data for classifying morals, we have used distant supervision to collect a large set of argumentative sentences automatically annotated for moral foundations. Training a BERT-based classifier on the data yielded state-of-the-art results. We have integrated the classifier with functionalities of Project Debater to tune arguments tuned towards specific morals. According to our user study, arguments with morals relevant to liberals are more effective than arguments without any control of morals. Also, conservatives were more affected by moral arguments. Our results demonstrate the feasibility of generating morally framed arguments. By focusing on shared morals, we believe that a respective system helps bridge disagreement of opposing audiences in practice.
\section*{Acknowledgments}

This work was funded by the Deutsche Forschungsgemeinschaft (DFG, German Research Foundation): TRR 318/1 2021 - 438445824. We would also like to thank the reviewers and the participants who took part anonymously in our user study.
\section{Ethical Statement}

The responsible administrative board of Paderborn University formally approved our external study. We did not gather any personal information about the participants that could connect their ideologies to their identity. Accordingly, no sensitive information was sent via the Project Debater API. Also, we ensured that they get paid more than the minimum wage in the U.S., namely 75\$ for a workload of 3 to 3.5 hours. Once more, we would like to indicate here that the results of this paper should be considered preliminary due to the limited number of surveyed users. Budget constraints did not allow to increase this number.

Working on technologies that aim to persuade an audience raises ethical concerns. Using knowledge about a user's morals is a critical act, and if done, it should be clearly communicated to the users. Similarly, generating arguments aiming to convince the audience could be thought of as a manipulation attempt. To avoid manipulation, any technology aiming at convincing the audience should be transparent about it by keeping the user informed of how their information is being used. Since we rely on the arguments provided by the Project Debater API, we ultimately cannot control their content, but our internal quality assessment study did not raise any notable concerns on the information contained.

As mentioned, the long-term goal of our envisioned system is not to use the audience's morals to generate arguments that convince them. Rather, given two disagreeing parties, the goal is to generate a wider spectrum of arguments covering relevant morals for both parties. We have argued in this paper that such morally rich arguments are more effective and could better achieve agreement between the disagreeing parties.

\bibliography{acl22-moral-debater-lit}
\bibliographystyle{acl_natbib}

\appendix
\section{The follow-up questionnaire}
This section details the follow-up questionnaire we carried to investigate users' judgments of effectiveness in both challenging and empowering arguments. In the questionnaire, we asked our six annotators the following four questions regarding the challenging arguments:
\begin{itemize}
	\item {\bf Your views}: When arguments contested your stance on the topic, which of the following arguments did you see as more effective:
		\begin{enumerate}
			\item Arguments that matched your views
			\item Arguments that challenged your views 
			\item Neither of those was important
		\end{enumerate}
	\item {\bf Your knowledge}: When arguments contested your stance on the topic, which of the following arguments did you see as more effective:
		\begin{enumerate}
			\item Arguments based on views you already knew about
			\item Arguments that introduce views you were not familiar with
			\item Neither of those was important
		\end{enumerate}

	\item {\bf Others' views}: When arguments contested your stance on the topic, which of the following arguments did you see as more effective:
		\begin{enumerate}
			\item Arguments you saw as particularly convincing to people that share your views
			\item Arguments you saw as particularly convincing to people that rather oppose your views
			\item Neither of those was important		
		\end{enumerate}

	\item {\bf Effectiveness}: When arguments contested your stance on the topic, which of the above three was most important for you to judge about effectiveness:
		\begin{enumerate}
			\item Your views
			\item Your knowledge
			\item Others' views
		\end{enumerate}
\end{itemize}

\begin{table*}[t!]%
	\centering%
	\small
	\renewcommand{\arraystretch}{1}
	\setlength{\tabcolsep}{3pt}%
	\begin{tabular}{llrrrrrrrrr}
		\toprule
		& & \multicolumn{3}{c}{\bf Conservatives} & \multicolumn{3}{c}{\bf Liberals} & \multicolumn{3}{c}{\bf ALL}  \\
		\cmidrule(l@{2pt}r@{2pt}){3-5} \cmidrule(l@{2pt}r@{2pt}){6-8} \cmidrule(l@{2pt}r@{2pt}){9-11}
		& & \bf Empow. & \bf Chall. & \bf All & \bf Empow. & \bf Chall. & \bf All & \bf Empow. & \bf Chall. & \bf All \\
		\midrule
		\bf Knowledge & Know about 	 & 33.3\% & 0.0\% & 16.7\% & 33.3\% & 0.0\% & 16.7\% & 33.3\% & 0.0\% & 16.7\% \\ 
					  & Not familiar & \bf 66.7\% & \bf 66.7\% & \bf 66.7\% & \bf 66.7\% & \bf 100.0\% & \bf 83.3\% & \bf 66.7\% & \bf 83.3\% & \bf 75.0\% \\ 
					  & Neither 	 & 0.0\% & 33.3\% & 16.7\% & 0.0\% & 0.0\% & 0.0\% & 0.0\% & 16.7\% & 8.3\% \\		
		\midrule
		\bf Own views  & Matched     & \bf 33.3\% & 0.0\%       & 16.7\%     & \bf 66.7\% & 33.3\%     & \bf 50.0\% & \bf 50.0\%   & 16.7\% & 33.3\% \\ 
						& Challenging & \bf 33.3\% & \bf 100.0\% & \bf 66.7\% & 33.7\%     & 0.0\%      & 16.7\%     & 33.3\% & \bf  50.0\%  & \bf 41.7\% \\ 
						& Neither     & \bf 33.3\% & 0.0\%       & 16.7\%     & 0.0\%      & \bf 66.7\% & 33.3\%     & 16.7\%       & 33.3\% & 25.0\% \\
		\midrule
		\bf Others' views & Share view  & 0.0\%      & 0.0\% & 0.0\%       & 33.3\%     & 0.0\%       & 16.7\%     & 16.7\%    & 0.0\% & 8.3\% \\ 
						& Oppose view & \bf 66.7\% & 33.3\% & \bf 50.0\% & \bf 66.7\% & \bf 100.0\% & \bf 83.3\% &\bf 66.7\% &\bf 66.7\% & \bf 66.7\% \\ 
						& Neither 	  & 33.3\%     &\bf 66.7\% & \bf 50.0\% & 0.0\%   & 0.0\%       & 0.0\%      & 16.7\%    & 33.3\% & 25.0\% \\
		
		\midrule
		\bf Effectiveness & Knowledge & \bf 66.7\% & \bf 66.7\% & \bf 66.7\% & \bf 100.0\% & \bf 66.7\% & \bf 83.3\% & \bf 83.3\% & \bf66.7\% & \bf75.0\% \\ 
						  & Views     & 33.3\% & 33.3\% & 33.3\% & 0.0\% & 33.3\% & 16.7\% & 16.7\% & 33.3\% & 25.0\% \\ 
						  & Neither   & 0.0\% &  0.0\% &  0.0\% &  0.0\% &  0.0\% &  0.0\% &  0.0\% &  0.0\% &  0.0\% \\
		\bottomrule
	\end{tabular}%
	\caption{Distribution of preferences (options) selected by the liberal and conservative annotators for each of the four asked questions for both the empowering and challenging cases.}
	\label{}%
\end{table*}

Similarly, we ask the same questions in the case of empowering arguments. Table 9 summarizes the results of the four asked questions in both empowering and challenging arguments. In general, among the three aspects, the knowledge aspect was the most relevant one to the majority of the annotators (last row). When the annotators were asked about the their own views (first row), conservatives favored mostly arguments that challenged their own views, while liberals favored mostly empowering arguments that matched their views. Regarding others' views, most conservatives and liberals valued empowering arguments that were particularly convincing to people that oppose their views. Nevertheless, we acknowledge the limited reliability of such self-assessment of one's moral judgments due to the complicated cognitive mechanisms behind it \cite{pizarro:2000}.
\newpage

\section{Example Generated Arguments}

We present in Table 7-9 a sample of generated arguments on three topics focusing on both individualizing and binding morals, as well as uncontrolled.

\begin{table*}[t!]%
	\centering%
	\small
	\renewcommand{\arraystretch}{1.0}
	\setlength{\tabcolsep}{2.5pt}%	
	\begin{tabular}{p{0.98\textwidth}}
		\toprule
		{\bf Topic: Globalization}  \\
		\midrule
		{\bf Binding argument:} The crowd raised four issues, explaining its views. The first claim is that globalization is reducing the importance of nation-states. The next issue will show how Globalization and structural forces aggravate poverty. In addition, we will hear about pollution and Culture. 
		
		Starting with nation states.  Globalization is reducing the importance of nation-states.   It has fueled the rise of transnational corporations, and their power has vaulted to the point where they can now rival many nation states.   This trend may threaten national identity because globalization undermines the importance of being a citizen of a particular country.  According to Ramonet, globalization and ultra-liberalism threaten the sovereignty of national states.  Some scholars have argued that increasing globalization has actually led to a contemporary form of imperialism, in which economic and cultural norms and discourses are imposed by dominant nations on weaker nations.  
		
		Turning to poverty.  Critics of globalization say that it disadvantages poorer countries.   Globalization and structural forces aggravate poverty and continue to push individuals to the margins of society.   Academic contributors to The Routledge Handbook of Poverty in the United States postulate that new and extreme forms of poverty have emerged in the U.S. as a result of neoliberal structural adjustment policies and globalization, which have rendered economically marginalized communities as destitute "surplus populations" in need of control and punishment.  
		
		Pollution was also mentioned.  Globalization creates risks that concern people from all different classes; for example, radioactivity, pollution, and even unemployment.   As International commerce develops new trade routes, markets and products it facilitates the spread of invasive species.  
		
		Lastly, Culture.  Globalization has destabilized previously immutable social institutions, shifting cultural value away from old traditions to new more individualistic and market friendly ideas.   It is often said to have a negative effect on the world's cultural diversity.   Cultural and geographical dimensions of transformational leadership become blurred as globalization renders ethnically specific collectivist and individualistic effects of organizational behavior obsolete in a more diversified workplace.  \\
		\midrule
		{\bf Individualizing argument:} 
		The crowd raised four issues, explaining its views. The first claim is that Globalization on its own cannot end gender inequality. In addition, we will hear about harm, economy and processes. 
		
		Starting with gender inequality.  There are various studies available that depict globalization as a hindrance toward gender inequality.   Globalization on its own cannot end gender inequality.  
		
		Turning to harm.  Some critics of globalization argue that it harms the diversity of cultures.   Globalization is a threat to culture and religion, and it harms indigenous people groups while multinational corporations profit from it.   It has been criticized for benefiting those who are already large and in power at the risk and growing vulnerability of the countries' indigenous population.   As of recently, it has been argued that globalization poses a threat to the CHS way of thinking because it often leads to the dissolution of distinct states.  
		
		Economy was also mentioned. According to many left-wing greens, economic globalization is considered a threat to well-being, which will replace natural environments and local cultures with a single trade economy, termed the global economic monoculture.  Globalisation is the cause of slow growth, unemployment and excessive debt in Western economies.   It makes people poor by taking away money from people who formerly had it.   Globalization does not work for all the economies that it affects, and that it does not always deliver the economic growth that is expected of it.   It is pretty much making all poor countries richer with the exception of various african countries because of their continious state of war between the local tribes.  
		
		Lastly, processes.  Processes of globalization present humankind with many issues that are considered problematic in at least one culture or society, and often multiple societies.   The books suggest that contemporary processes of globalization and mediatization have contributed to materially abstracting relations between people, with major consequences for how we live our lives.   On the other side, the process of globalization has generated universal disenchantment.  
		
		 \\
		\midrule
		{\bf Uncontrolled argument:} 
		The crowd raised four issues, explaining its views. The first claim is that globalisation creates economic and cultural imbalances in developing nations. The next issue will show how globalization is reducing the importance of nation-states. And the third point is that globalization is a threat. In addition, we will hear about processes. 
		
		Starting with economy.  Globalization does not work for all the economies that it affects, and that it does not always deliver the economic growth that is expected of it.   Globalisation and neoliberalism have exacerbated already unequal economic relations.   Although globalization takes similar steps in most countries, scholars such as Hodge claim that it might not be effective to certain countries and that globalization has actually moved some countries backward instead of developing them.  
		
		Turning to nation states.  Globalization is reducing the importance of nation-states.   It engenders conflicts within and between nations over domestic norms and social institutions.   This trend may threaten national identity because globalization undermines the importance of being a citizen of a particular country.  According to Ramonet, globalization and ultra-liberalism threaten the sovereignty of national states. 
		
		Threats was also mentioned.  Case studies of Thailand and the Arab nations' view of globalization show that globalization is a threat to culture and religion, and it harms indigenous people groups while multinational corporations profit from it.   As of recently, it has been argued that globalization poses a threat to the CHS way of thinking because it often leads to the dissolution of distinct states.  
		
		Lastly, processes.  The combined processes of industrialization and globalization have disrupted longstanding livelihoods and systems of production, forcing many families to focus more on income-generating activities than on subsistence practices.   Academic contributors to The Routledge Handbook of Poverty in the United States postulate that new and extreme forms of poverty have emerged in the U.S. as a result of neoliberal structural adjustment policies and globalization, which have rendered economically marginalized communities as destitute "surplus populations" in need of control and punishment.  
		
		 \\
		\bottomrule
	\end{tabular} 
	\caption{
		Example generated arguments against \textit{Globalization} for different focused morals.
	}
	\label{table-example-generated-arguments-full}
\end{table*}

\begin{table*}[t!]%
	\centering%
	\small
	\renewcommand{\arraystretch}{1.0}
	\setlength{\tabcolsep}{2.5pt}%	
	\begin{tabular}{p{0.98\textwidth}}
		\toprule
		{\bf Topic: Affirmative Actions}  \\
		\midrule
		{\bf Binding argument:} 
		Law, students, policy and women are the four issues the crowd elaborated on. 
		
		Let's explore the issue of law. The court found that the University of Michigan's Law School's affirmative action admission policies were promoting diversity within its school. Liberals have also argued for affirmative action to increase the representation of women and minorities among law students and law faculty. Following the enactment of Civil Rights laws in the 1960s, all educational institutions in the United States that receive federal funding have undertaken affirmative action to increase their racial diversity. 
		
		The next issue is students. Another controversial decision of the Rehnquist court in 2003 was Grutter v. Bollinger which upheld affirmative action. It was in essence an affirmative action scheme to assist geographically disadvantaged students to gain tertiary education. The obvious solution to all the Affirmative Action controversy is to offer full financial assistance to all university students with need, regardless of race. 
		
		Moving on to policy. This executive order created a National Women's Business Enterprise Policy and required government agencies to take affirmative action in support of women's business enterprises. Following the riots, the Malaysian Government introduced affirmative action policies to help the Bumiputera to achieve a higher economic quality of life than the Chinese. Ethiopian government policy has supported affirmative action for women since its inception in 1994. The ACLU's official position statements, as of January 2012, included the following policies: Affirmative action - The ACLU supports affirmative action. 
		
		The last issue mentioned was women. Hunter is a supporter of affirmative action for women. He is highly in favor of affirmative action and supports setting aside funds for women and minorities. According to a poll taken by USA Today in 2005, the majority of Americans support affirmative action for women, while views on minority groups were more split. It support affirmative action for women. Affirmative Action and Impact Sourcing Tata BSS is an avid supporter of Affirmative Action and hence employs huge number of SC/ST community people in their operations thereby making a huge impact on the society.
		  \\
		\midrule
		{\bf Individualizing argument:} 
		Employment, discrimination, universities and policy are the four issues the crowd elaborated on. 
		
		Let's explore the issue of employment. Affirmative action legislation has led to substantial improvements in the employment of minorities and women. They also require Federal contractors and subcontractors to take affirmative action to ensure equal employment opportunity in their employment processes. Historically and internationally, support for affirmative action has sought to achieve goals such as bridging inequalities in employment and pay, increasing access to education, promoting diversity, and redressing apparent past wrongs, harms, or hindrance. 
		
		The next issue is discrimination. The affirmative action program is designed to remedy the effects of past discrimination. In 2004, he reiterated his support, "I support affirmative action programs, including in appropriate instances consideration of race and gender in government contracting decisions, when the affirmative action program is designed to remedy the effects of past discrimination.". Alongside the prohibition against unfair discrimination, affirmative action is the second cornerstone of the EEA. A common approach to remedying discrimination is affirmative action. 
		
		Moving on to universities. The National Conference of State Legislatures held in Washington D.C. stated in a 2014 overview that many supporters for affirmative action argue that policies stemming from affirmative action help to open doors for historically excluded groups in workplace settings and higher education. Race-based affirmative action was necessary to achieve diversity and its educational benefits. Affirmative action provides some disadvantaged youth with increased chances of attending top-tier university. Action is thus more important than abstract knowledge. 
		
		The last issue mentioned was policy. According to its website, the NCP is committed to ideals of social justice, expressing support for affirmative action policies for the downtrodden members of society and for ensuring equal opportunities for all. Such legislation and affirmative action policies have been critical to bringing changes in societal attitudes. The school applies an affirmative action policy to ensure marginalized students from hardship semi-arid areas are given a chance. Affirmative Action Policy is to ensure the peace and stability in the pluralist society of Malaysia. It was found to address Malay grievances.
		
		\\
		\midrule
		{\bf Uncontrolled argument:} 
		
		Discrimination, effectiveness, the supreme court and women are the four issues the crowd elaborated on. 
		
		Let's explore the issue of discrimination. 1961: Executive Order 10925: Required government contractors to "take affirmative action" to ensure non-discriminatory employment practices. It supports affirmative action as necessary in the fight for equality and it opposes all forms of racism and religious bigotry. Right to affirmative action All federal employers or federal contractors are required to take affirmative action to help counteract the effects of historical discrimination. CAMERA argues the Law of Return is justified under the Convention on the Elimination of All Forms of Racial Discrimination Article I, which CAMERA argues allows for affirmative action, because of the discrimination Jews faced during the Holocaust. 
		
		The next issue is effectiveness. In philosophy, Action is effective will. Collective action is the most effective means of preventing potential state and non-state aggressors from threatening other states. Such action is effective in spreading our message. 
		
		Moving on to the supreme court. Utter v. Bollinger: The Supreme Court of the United States upholds affirmative action in university admissions. The court found that the University of Michigan's Law School's affirmative action admission policies were promoting diversity within its school. Importantly, though, it ruled that a university was entitled to "no deference" on its judgment that race-based affirmative action was necessary to achieve diversity and its educational benefits. Another controversial decision of the Rehnquist court in 2003 was Grutter v. Bollinger which upheld affirmative action. A prominent advocate of affirmative action, he played a leading role in the twin Supreme Court cases-Grutter v Bollinger and Gratz v Bollinger-that upheld and clarified the importance of diversity as a compelling justification for affirmative action in higher education. 
		
		The last issue mentioned was women. Affirmative action programs to women in an effort to end gender discrimination. Gender-affirmative practices recognize and support an individual's unique gender self-identification and expression. Wesolowski supported affirmative action programs for Saskatoon Catholic teachers in 2001, arguing that past policies had discriminated against women and aboriginals. For women business owners, affirmative action programs are considered legislation that encourage government agencies and contractors to do business with women-owned businesses.	
		
		\\
		\bottomrule
	\end{tabular} 
	\caption{
		Example generated arguments supporting \textit{Affirmative Actions} for different focused morals.
	}
	\label{table-example-generated-arguments-full}
\end{table*}

\begin{table*}[t!]%
	\centering%
	\small
	\renewcommand{\arraystretch}{1.0}
	\setlength{\tabcolsep}{2.5pt}%	
	\begin{tabular}{p{0.98\textwidth}}
		\toprule
		{\bf Topic: Legalizing Prostitution}  \\
		\midrule
		{\bf Binding argument:} 
		
		These are the four issues the crowd addressed: the netherlands, adulthood, government and drugs. 
		
		Starting with the netherlands. Prostitution is legal and regulated in the Netherlands. In Switzerland, it has been legal since 1942. As a result of such views on prostitution, countries such as Germany, the Netherlands and New Zealand have fully legalized prostitution. 
		
		Adulthood was also mentioned. Prostitution of adults is legal in 10 rural counties in Nevada. Prostitution is a consensual sex act between adults and a victimless crime, thus the government should not prohibit this practice. The impugned Criminal Code provisions, by criminalizing many of the activities surrounding prostitution, adversely affect a great number of women. Prostitution was legal, as long as the women paid their license fees. It should not be illegal because the customers are satisfied. 
		
		Turning to government. Prostitution should be decriminalized, and as a libertarian insists the sex trade should not be regulated by the government. According to data from the Office for National Statistics, it contributed 5.3 billion to the UK economy in 2009. This, the sex work perspective asserts, will allow prostitution to be regulated by governments and business codes, protect sex trade workers, and improve the ability to prosecute people who hurt them. In 2005 the government also drafted the Sex Regulation Act which sought to further legalise and regulate prostitution. According to a Portuguese Government spokesperson, "The Government's opinion was that prostitution was not a crime. 
		
		The last issue mentioned was drugs. Mick Jagger has called for all drugs to be legalised on the Isle of Man. The Act legalised prostitution and put the women involved under police and medical control. In 2012, newly elected Guatemalan president Otto Perez Molina argued that all drugs should be legalized while attending the United Nations. Because of this, the Austrian Federal Ministry of the Interior wants to transform illegal prostitution into legal regulated prostitution.	
		
		\\
		\midrule
		{\bf Individualizing argument:} 
		These are the four issues the crowd addressed: regulation, gambling, rape and crime. 
		
		Starting with regulation. According to proponents of regulation, prostitution should be considered a legitimate activity, which must be recognized and regulated, in order to protect the workers' rights and to prevent abuse. It should be legalised so it could be controlled safety. Maxine is a strong supporter of legalised brothels, more regulations around employment in the sex industry and increased sentences for perpetrators of violent crime. Barbara Brents and Kathryn Hausbeck state in their study that the legalization of prostitution in Nevada's brothels allows for improved regulation and protection for both businesses and workers. 
		
		Gambling was also mentioned. D'Amato believed that legalized gambling would be good for both the city and for those with businesses related to the entertainment industry. Maher favors ending corporate welfare and federal funding of non-profits as well as the legalization of gambling, prostitution, and cannabis. 
		
		Turning to rape. Kimberly Kay Hoang, assistant professor of sociology at the University of Chicago, who conducted a 2011 study of prostitutes in Ho Chi Minh City is quoted as saying "Legalising prostitution would also reduce violence and sex crimes such as rape and sexual violence. Legalising prostitution would also reduce violence and sex crimes such as rape and sexual violence. In 2006, the National Assembly legalized abortion care in cases of rape, as women regularly faced sexual violence, rape, and gang rape in the war-ravaged country. Legalizing brothels would make prostitution safer for women because it would allow prostitution to take place indoors. 
		
		The last issue mentioned was crime. Proposition K would decriminalize prostitution, enforce laws against crimes on sex workers, and disclose all investigations and prosecutions of violent crimes against sex workers. A May 1990 Australian Institute of Criminology report recommended that prostitution not be a criminal offence, since the laws were ineffective and endangered sex workers. In 1908, the king passed laws to legalize prostitution and help sex workers get medical care.			
		
		\\
		\midrule
		{\bf Uncontrolled argument:} 
		
		These are the four issues the crowd addressed: offices, sex workers, drugs and rural. 
		
		Starting with offices. Citizens' Assembly's Vanadzor Office president, Artur Sakunts, called for prostitution to be legalised and regulated. Prostitution in Hungary has been legalized and regulated by the government since 1999. In several countries, lotteries are legalized by the governments themselves. In response to the 1995 Federal-Provincial-Territorial Working Group on Prostitution report "Dealing with Prostitution in Canada," Toronto's Board of Health advocated decriminalisation in 1995, with the City taking the responsibility of regulating the industry. 
		
		Sex workers was also mentioned. The sex workers organisation "Guyana Sex Worker Coalition" and several NGOs called for prostitution to be legalised and regularization of sex work. Some sex-positive feminists believe that women and men can have positive experiences as sex workers and that where it is illegal, prostitution should be decriminalized. Since the mid-1970s, sex workers across the world have organised, demanding the decriminalisation of prostitution, equal protection under the law, improved working conditions, the right to pay taxes, travel and receive social benefits such as pensions. The sex work perspective maintains that prostitution is a legitimate form of work for women faced with the option of other bad jobs, therefore women ought to have the right to work in the sex trade free of prosecution or the fear of it. 
		
		Turning to drugs. If they did, prostitution and drugs would be legal. The Act legalised prostitution and put the women involved under police and medical control. In 2012, newly elected Guatemalan president Otto Perez Molina argued that all drugs should be legalized while attending the United Nations. He has studied the effects of drug criminalization for 15 years, and argues that all drugs should be legalized. 
		
		The last issue mentioned was rural. Prostitution of adults is legal in 11 rural counties in Nevada. Mayor Goodman supports legalizing prostitution in the city's downtown area as a revenue generator and tool for revitalization, although a majority of Nevadans polled in 2003 opposed the mayor's position.
		
		\\
		\bottomrule
	\end{tabular} 
	\caption{
		Example generated arguments supporting \textit{Legalizing Prostitution} for different focused morals.
	}
	\label{table-example-generated-arguments-full}
\end{table*}
\end{document}